\documentclass[runningheads]{llncs}
\usepackage[T1]{fontenc}
\usepackage{hyperref}
\usepackage{amssymb}
\usepackage{multirow}
\usepackage{booktabs}
\usepackage{graphicx}
\begin{document}
%

\title{ Impact of Iris Pigmentation on Performance Bias in Visible Iris Verification Systems: A Comparative Study }
%
\author{Geetanjali Sharma\inst{1} \and
Abhishek Tandon\inst {1} \and
Gaurav Jaswal\inst{2} \and
Aditya Nigam\inst{1} \and
Raghavendra Ramachandra \inst{3}}
\authorrunning{Geetanjali Sharma et al.}
%
\institute{Indian Institute of Technology Mandi, India \and
Technology Innovation Hub, Indian Institute of Technology Mandi, India \and
Norwegian University of Science and Technology (NTNU), Norway
\\}
\titlerunning{Impact of Iris Pigmentation on Performance Bias in Visible Iris Verification Systems: A Comparative Study}
\maketitle              
\begin{abstract}


Iris recognition technology plays a critical role in biometric identification systems, but their performance can be affected by variations in iris pigmentation. In this work, we investigate the impact of iris pigmentation on the efficacy of biometric recognition systems, focusing on a comparative analysis of blue and dark irises. Data sets were collected using multiple devices, including P1, P2, and P3 smartphones \cite{8698587}, to assess the robustness of the systems in different capture environments \cite{8272762}. Both traditional machine learning techniques and deep learning models were used, namely Open-Iris, ViT-b, and ResNet50, to evaluate performance metrics such as Equal Error Rate (EER) and True Match Rate (TMR). Our results indicate that iris recognition systems generally exhibit higher accuracy for blue irises compared to dark irises. Furthermore, we examined the generalization capabilities of these systems across different iris colors and devices, finding that while training on diverse datasets enhances recognition performance, the degree of improvement is contingent on the specific model and device used. Our analysis also identifies inherent biases in recognition performance related to iris color and cross-device variability. These findings underscore the need for more inclusive dataset collection and model refinement to reduce bias and promote equitable biometric recognition across varying iris pigmentation and device configurations.

\end{abstract}
\section{Introduction}
Iris recognition systems have become one of the most reliable and secure forms of biometric identification \cite{raghavendra2014presentation}\cite{raghavendra2015exploring}\cite{yuan2020defense} due to unique and stable patterns in the human iris. These systems leverage high-resolution imaging and advanced algorithms \cite{ahmad2019thirdeye}\cite{ren2023multiscale} to capture, extract, and match intricate iris features, offering unmatched accuracy in identity verification. Despite their robustness, the fairness \cite{kotwal2022fairness} of iris recognition systems has become a critical area of research. Variations in iris color, texture, and pupil dilation due to demographic differences can introduce biases, affecting the system's performance across different populations. For instance, factors such as age, ethnicity, and even environmental conditions can lead to discrepancies in recognition accuracy, raising concerns about equitable treatment for all users. Specifically, the risk of bias based on demographic factors such as age, race, or gender can result in uneven performance across different groups. This performance disparity, often quantified through differential error rates, highlights the need to address these biases to ensure that biometric systems provide fair and equitable outcomes for all users
This paper delves into the technical foundations of iris recognition systems, explores the inherent biases that can affect their performance, and discusses strategies to enhance fairness in these systems. Through comprehensive analysis and experimentation, we aim to contribute to the ongoing efforts to create more equitable biometric technologies.
\subsection{Motivation and contribution}
Fairness in biometric systems specifically concerns differences in outcomes linked to demographic traits\cite{rathgeb2022demographic}\cite{andrus2022demographic}\cite{valdivia2023there}, such as age, gender, race, or iris color. However, these demographic variations\cite{valdivia2023there} may overlap with other individual characteristics, leading to complex biases.
\begin{figure}[htbp] 
\centering
\includegraphics[scale=0.4]{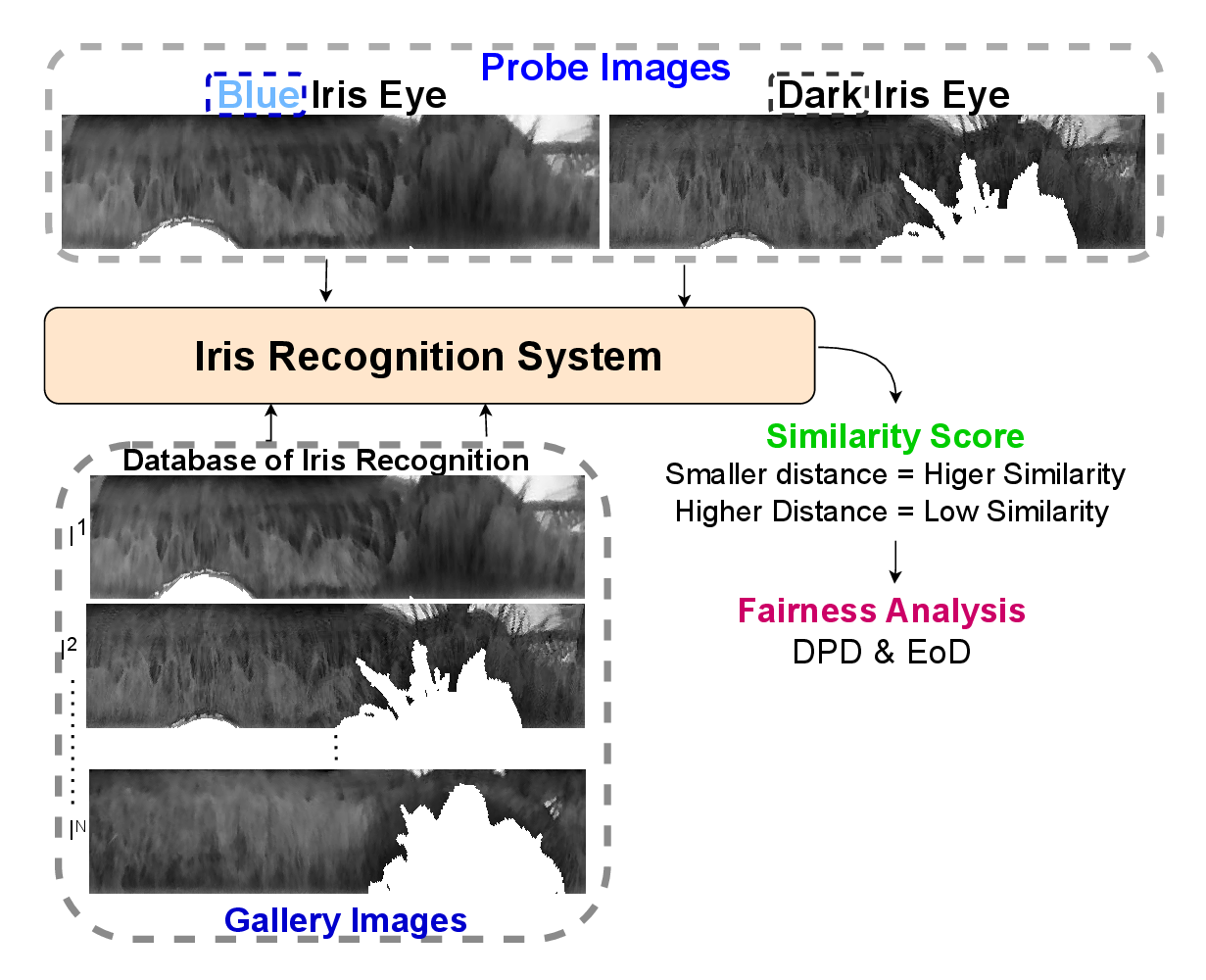}
\caption{Illustrates an iris recognition system comparing blue and dark irises, using similarity scores and analyzing fairness with DPD and EOD metrics to address demographic biases.}
\label{figure:motivation}
\end{figure}

These differences in iris pigmentation \cite{edwards2016iris} might affect the system's ability to accurately match or verify identity, leading to higher false rejection rates for some groups. Highlights the importance of developing algorithms \cite{rathgeb2022demographic}\cite{kotwal2022fairness} that account for these variations to ensure that biometric systems\cite{jalilian2022deep}\cite{yang2021dualsanet} perform consistently across all demographics, thereby avoiding unintended biases in high-stakes applications. The primary aim of fairness in iris biometrics\cite{edwards2016iris} is to identify and understand biases in systems that determine or validate individual identities or characteristics, such as iris color. In our study, we are motivated to investigate and mitigate biases in iris biometric systems to ensure fairness and improve the accuracy and reliability of these systems, particularly in relation to iris pigmentation color as shown in Figure \ref{figure:motivation}. This research aims to address the growing concerns of fairness in biometric systems by examining how different pigmentation levels, such as blue versus dark irises, impact system performance. To the best of our knowledge, this is the first study to comprehensively explore these effects across multiple devices and recognition models, highlighting the importance of unbiased and accurate biometric recognition. Our work aims to tackle the rising challenge of ensuring fairness in biometric systems, particularly focusing on how biases related to iris pigmentation can impact the accuracy and reliability of recognition models. In particular, we introduce the following critical questions:
\begin{itemize}
    \item \textbf{Q1.}Does iris pigmentation  (Blue Iris (BI) vs. Dark Iris (DI) color) impact the accuracy of the biometric system ?\\
    \item \textbf{Q2.}Can biometric systems effectively generalize across varying iris colors, specifically between blue and dark iris ?
\end{itemize}
In addressing the research questions posed above, this study makes the following key contributions, including being the first to investigate the bias of visible iris recognition systems for blue and dark-colored irises. 

\begin{itemize}
    \item To the best of author's knowledge, this is the first detailed study examining how visible blue vs dark iris pigmentation impacts the performance of iris recognition systems.
    \item A detailed experimental study are presented to benchmark the performance of the traditional and deep learning-based iris recognition systems to provide insights into the generalization capability of biometric systems across different smartphone devices and iris pigmentation colors ( Blue vs Dark).
    \item A fairness analysis is conducted using Demographic Parity Difference (DPD) and Equalized Odds Difference (EoD) metrics to assess biases in the recognition systems.
\end{itemize}
The structure of the paper is as follows: Section \ref{sec:2_LR} reviews related work in iris recognition. Section \ref{sec:3_methodology} explores various methods employed in iris recognition, while Section \ref{sec:4_Experiments_results} presents a quantitative and qualitative analysis of these systems, examining their impact on iris pigmentation (dark versus blue). Finally, Section \ref{sec:5_conclusion} concludes the paper.

\section{Literature review}
\label{sec:2_LR}
In this section, we first review the related work on iris recognition, focusing on both Near-Infrared (NIR) and Visible Spectrum (VIS) imaging techniques. We include an analysis of contact and contactless iris recognition, with images captured using smartphones and DSLR devices. Additionally, we also examine advancements in segmentation and recognition performance, while also assessing potential biases in VIS-based systems related to iris pigmentation.

Iris Recognition is a leading innovation in biometric recognition, offering unparalleled accuracy for secure identity verification. Whether used for crossing international borders, safeguarding financial data, or securing smartphones, recognition plays an important role. Using the unique patterns of the iris provides accuracy and reliability that is better than most other biometric systems. First Daugman’s seminal work \cite{daugman2009iris} established the foundation for modern iris recognition technology, introducing an efficient algorithm based on Gabor wavelets for encoding and matching iris patterns, which has since become a fundamental principle in biometric security systems. 

Near Infrared (NIR) iris recognition has been extensively studied for its effectiveness in capturing detailed iris patterns, even in low-light conditions. NIR imaging penetrates through the corneal surface, revealing the unique texture of the iris, which is crucial for accurate biometric verification \cite{daugman2009iris}. 
Despite progress in deep learning for iris segmentation, challenges still exist in non-cooperative environments, especially with NIR images, where accurate segmentation is essential for tasks like normalization and recognition \cite{wang2021nir} \cite{li2021robust}. In \cite{wang2021nir}, segmenting NIR iris images when the person is wearing contactless glasses is difficult due to reflections and obstructions from the glasses, which block details of the iris. Also, changing light conditions can cause uneven contrast, making it harder to accurately detect the boundaries.
In visible spectrum iris recognition, capturing clear patterns in dark irises can be challenging due to melanin pigmentation and collagen fibrils, which obscure texture details under visible light. In \cite{raja2015iris} study introduces a novel method using white LED light to enhance image quality and improve verification performance, demonstrating promising results with a new dataset of dark irises and high accuracy in real-world applications.  Another work \cite{gong2012optimized}  explores iris imaging across 12 wavelengths from 420 to 940 nm, aiming to identify the optimal wavelength for recognizing heavily pigmented irises, and the optimal wavelength band was determined based on texture quality and performance metrics. The specific wavelength band enhances recognition accuracy for heavily pigmented irises\cite{gong2012optimized}.

Iris biometrics, traditionally used in controlled settings, are now being adapted for mobile applications, such as smartphones, which pose unique challenges. In \cite{thavalengal2015evaluation} evaluates a dual-visible/NIR camera system with 4 Megapixel resolution for iris recognition in unconstrained environments, revealing issues with optical performance and resolution limits, but suggesting that enhanced optics could integrate biometric functions with standard front-camera features. The appearance of high-resolution smartphone cameras has greatly facilitated the capture of visible light (VIS) iris images \cite{raja2015smartphone}. These smartphones are valued for their convenience and their capability to capture detailed iris images without the need for specialized equipment. To address the limitations inherent in single-spectrum systems, Dual-Spectrum Recognition (DSR) cameras have been developed. These cameras capture both Near-Infrared (NIR) and VIS images, combining the strengths of each spectrum to improve recognition accuracy and robustness. 
Visible iris recognition systems are increasingly used for biometric identification, but they face significant bias-related challenges. In particular, NIR and VIS iris recognition systems often exhibit performance disparities based on iris pigmentation. Dark irises, characterized by high melanin levels, present difficulties in image capture and recognition due to reduced contrast and detail visibility compared to lighter irises. This issue has been highlighted in various studies, which emphasize that traditional systems are less effective for individuals with darker irises due to the limited texture visibility under visible light. Iris pigmentation refers to the coloration of the iris, which is influenced by melanin levels and structural variations. Pigmentation affects the visibility of iris patterns, which are critical for accurate biometric recognition. Figure \ref{figure:figure1} illustrates how different pigmentation levels can impact the appearance of iris textures. To the best of our knowledge, this is the first study to comprehensively investigate the impact of iris pigmentation, specifically focusing on the challenges associated with blue and dark irises in visible light systems. This research highlights the inherent biases in the latest iris recognition technologies and proposes novel approaches to improve performance across different pigmentation types.

\section{Methodology}
\label{sec:3_methodology}
\begin{figure}[htbp] 
\centering
\includegraphics[width=0.9\linewidth]{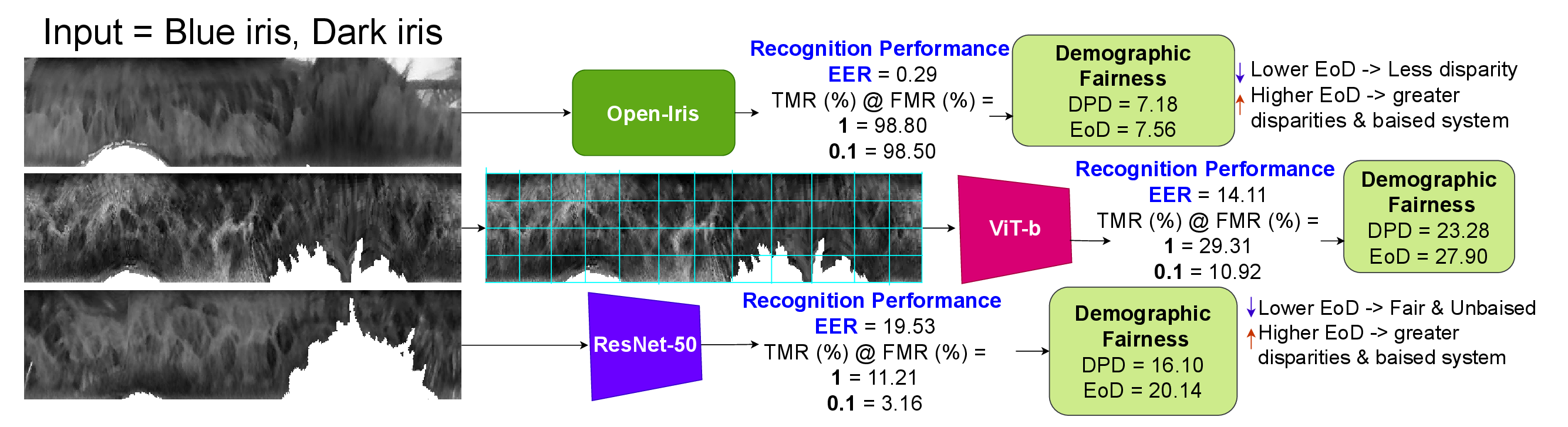}
\caption{Illustration of the iris recognition system's performance on a dataset with varying iris pigmentation (dark and blue irises) using three different models: Open-Iris, ViT-b, and ResNet-50. The study evaluates how iris pigmentation influences recognition accuracy and assesses fairness through Demographic Parity Difference (DPD) and Equalized Odds Difference (EoD) metrics to identify potential system biases and demographic disparities.} 
\label{figure:figure1}
\end{figure}

In this section, we describe the comprehensive approach taken to evaluate the impact of iris pigmentation and device variability on the performance of iris recognition systems. Our methodology involves several key steps: dataset collection and preparation, preprocessing, system evaluation, fairness analysis, and comparative analysis.
In dataset collection and preprocessing we used five distinct iris pigmentation (Blue and Dark iris color) image datasets to examine the performance of iris recognition systems across different pigmentation colors and device types.
To ensure consistency and quality, images were pre-processed to standardize size, enhance contrast, and reduce noise, while preserving essential pigmentation details.
In \textbf{preprocessing stage}, images go through preprocessing to optimize their quality for recognition algorithms. This process involved normalizing image sizes, adjusting contrast, and applying filters to minimize noise. The preprocessing aimed to mitigate variations in pigmentation visibility due to different imaging conditions while maintaining the integrity of the iris features. \\
\textbf{Iris Recognition System}: We evaluated three different iris recognition systems to assess their performance as shown in Figure \ref{figure:figure1}, Table \ref{table:table2}, and Table \ref{fig:table1-fig}.
\begin{itemize}
    \item \textbf{Open-Iris:}  
    Open-Iris is an iris recognition system using classical image processing techniques like edge detection and texture analysis, effective across various imaging conditions. We utilized Open-Iris as a feature extractor for iris recognition, particularly for images from smartphone devices with varying iris pigmentation.
    \item \textbf{ViT-B:} 
      The Vision Transformer (ViT-b) processes images as patches, capturing global patterns for effective iris recognition, especially with varying pigmentation. We fine-tuned the ViT base model on DI and BI datasets across different devices to enhance feature extraction for accurate iris recognition.

    \item \textbf{ResNet-50:}
   ResNet-50, a deep CNN with residual blocks, effectively extracts diverse iris pigmentation patterns. The extracted features are then fed into an SVM classifier to learn fine-grained details for improved iris recognition.  

\end{itemize}
\textbf{Evaluation Metrics:} 
The performance of the recognition systems is evaluated using Equal Error Rate (EER) and True Match Rate (TMR). EER measures when false acceptance and rejection rates are equal, indicating system accuracy. TMR assesses true matches at False Match Rate (FMR) thresholds of 1\% and 0.1\%, showing the system's ability to identify true positives under different false match tolerances.
\textbf{Fairness Analysis: }We conducted a fairness analysis to identify biases in iris recognition systems related to iris pigmentation (blue vs. dark). Key fairness metrics, Demographic Parity Difference (DPD) and Equalized Odds Difference (EoD), were calculated. DPD measures disparities in recognition rates across groups, while EoD assesses imbalances in error rates. These metrics provide a comprehensive view of system fairness, helping to identify and address potential biases.

\section{Experiments and Results}
\label{sec:4_Experiments_results}
In this section, we explore the impact of iris pigmentation\cite{edwards2016iris}, specifically comparing blue and dark irises, on the performance of visible iris recognition systems. We analyze how the color of the iris influences recognition accuracy by employing both traditional and deep learning-based approaches. To quantitatively evaluate the effectiveness of these iris recognition systems, we utilize the Open-Iris tool \cite{wldiris}, ResNet-50 \cite{he2016deep}, and ViT-b\cite{alexey2020image} system. This tool facilitates detailed analysis and comparison across different iris pigmentation types, offering insights into the robustness and accuracy of various recognition techniques.
Our experiment aims, to highlight recognition bias or performance disparities that may arise due to pigmentation differences.  
The performance of iris recognition is evaluated on both blue and dark irises, with images captured using various smartphone devices, including the iPhone 5S(we call it as P1), Nokia  Lumia 1020 (we call it as P2), and Samsung Galaxy S5 (we call it as P3). To ensure a comprehensive analysis, we assess the recognition performance across three distinct iris recognition systems, each employing different algorithms and methodologies. 

\subsection{Dataset}
\begin{figure}[htbp] 
\centering
\includegraphics[scale=0.25]{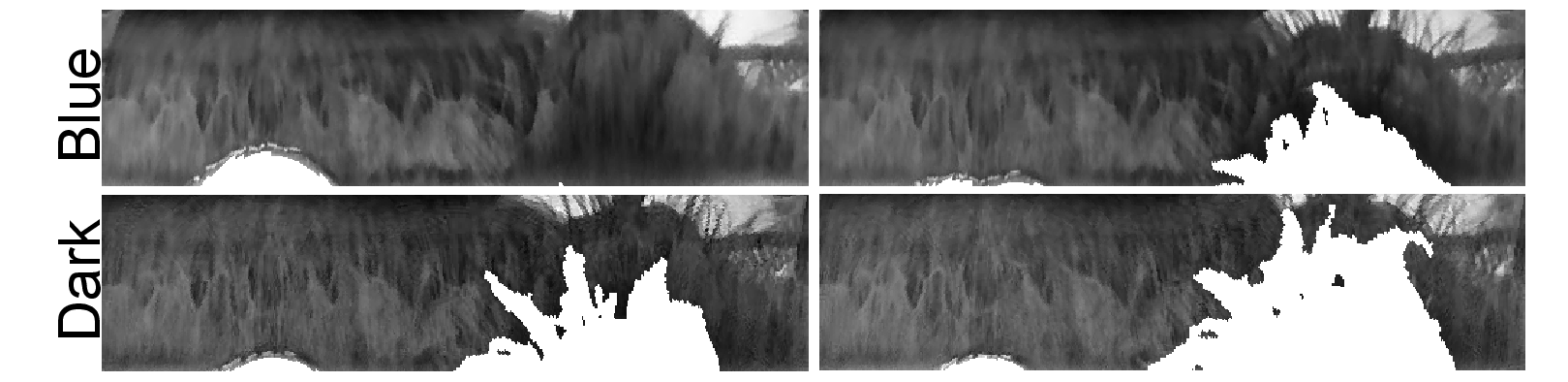}
\caption{Preprocessed blue and dark irises showing reduced pigmentation differences and noise removal (white pixels), highlighting uniformity for biometric analysis.}
\label{figure:dataset}
\end{figure}

Our dataset comprises five distinct iris datasets available from \cite{raja2015smartphone}, categorized by iris pigmentation and capture device collected from 58 unique subjects. Two of the datasets consist of blue irises, with images captured using a P1 and a P2 smartphone (sample iris images shown in Figure \ref{figure:dataset}. The remaining three datasets feature dark irises, collected using the rear cameras of different devices: P1, P2, and P3. In total, we have 58 subjects and We have equally divided the images into gallery and probe. This diverse collection allows us to thoroughly examine the impact of iris color and device variation on the performance of iris recognition systems.
\subsection{Performance Evaluation Protocol}
The performance of our system was evaluated by using processed iris strip images as input to deep learning models, which learned iris texture features to generate both genuine and imposter scores. Each subject contributed approximately five images, resulting in a dataset of 58 subjects. Genuine comparisons were obtained by matching each subject's 2 gallery images with their own 3 probe images, leading to 6 genuine matches per subject and a total of 348 genuine $(58*3*2=348)$ comparisons. Imposter comparisons were generated by matching each subject's 2 gallery images with the 3 probe images from all other 57 subjects, resulting in 342 imposter matches per subject $(57*3*2=342)$ and a total of 19,836 imposter $(342*58=19836)$ comparisons.

\subsection{Iris Recognition results and discussion}
Table \ref{table:table1} provides the performance metrics for the iris pigmentation (Blue Iris dataset using various iris recognition systems, including OpenIris \cite{wldiris}, ViT-B \cite{alexey2020image}, and ResNet-50 \cite{he2016deep}. The results are presented in terms of Equal Error Rate (EER\%), True Match Rate (TMR\%), and False Match Rate (FMR\%) at thresholds of 0.1\% and 0.01\%. This comprehensive evaluation offers insights into the impact of iris pigmentation on recognition accuracy across different systems.
\begin{table*}[htbp]
    \centering
    \caption{Iris recognition results for blue iris Dataset}
    \begin{tabular}{lcccc}
    \hline
    \multicolumn{5}{c}{\textbf{Iris Recognition (Blue Iris(BI) )}} \\ \hline
    \multirow{2}{*}{\textbf{Dataset}} & \multirow{2}{*}{\textbf{Models (\%)}} & \multirow{2}{*}{\textbf{EER (\%)}} & \multicolumn{2}{c}{\textbf{TMR (\%) @ FMR (\%) =}} \\ \cline{4-5}
     &  &  & \textbf{1 (\%)} & \textbf{0.1 (\%)} \\ \hline
    BI-P1 & Open-Iris & \textbf{0.24} & \textbf{99.71} & \textbf{98.85} \\ \hline
    BI-P1 & ViT-b & 8.94 & 66.95 & 42.24 \\ \hline
    BI-P1 & Resnet50 & 19.18 & 24.14 & 10.06 \\ \hline
    BI-P2 & Open-Iris & \textbf{0.82} & \textbf{99.43} & \textbf{91.67} \\ \hline
    BI-P2 & ViT-b & 14.11 & 54.60 &  29.31 \\ \hline
    BI-P2 & Resnet50 & 19.53 & 29.60 & 11.21 \\ \hline
    \end{tabular}    
    \label{table:table1}
\end{table*}
The results in Table \ref{table:table1} demonstrate the varying performance of different iris recognition models in blue iris data sets captured using devices P1 and P2. The Open-Iris model excels, achieving exceptionally low Equal Error Rates (EERs) of $0.24\%$ and $0.82\%$ for P1 and P2 datasets, respectively, along with impressively high True Match Rates (TMRs) exceeding $99\%$ at both$ 1\%$ and $0.1\%$ False Match Rate (FMR) thresholds. 
 \begin{figure}[!ht]
\centering
\includegraphics[scale=0.3]{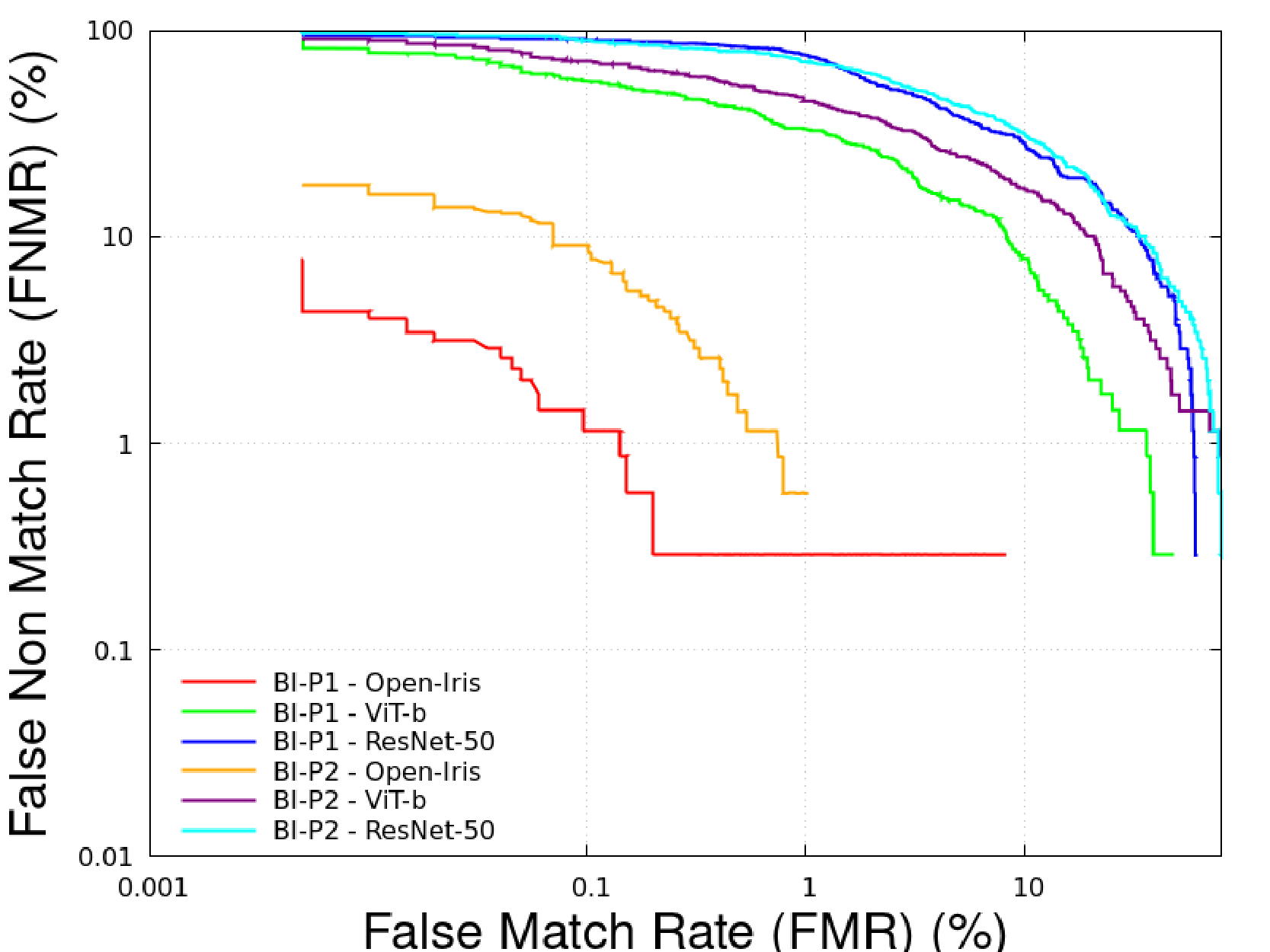}
\caption{ DET curve for comparative analysis among three different models. The X-axis indicates the false match rate and the y-axis indicates the false non-match rate of the Iris Pigmentation (Blue color) dataset.}
\label{fig:table1-fig}
\end{figure}
In contrast, the ViT-b and ResNet-50 models exhibit significantly weaker performance. The ViT-b model struggles particularly with higher EERs of $8.94\%$ and $14.11\%$, and substantially lower TMRs, especially at the more stringent $0.1\%$ FMR threshold, where it fails to provide any matches for the P1 dataset. Similarly, ResNet-50, while slightly better than ViT-b, still performs poorly, with EERs above $19\%$ and TMR's falling below $12\%$ across both datasets. These findings underscore the effectiveness of the Open-Iris model in accurately recognizing blue irises, whereas ViT-b and ResNet-50 are less reliable, particularly in more challenging scenarios, highlighting the critical role of model selection in iris recognition tasks.
\begin{table*}[htbp]
    \centering
    \caption{Iris recognition results for Dark iris Dataset}
    \begin{tabular}{lcccc}
    \hline
    \multicolumn{5}{c}{\textbf{Iris Recognition (Dark Iris (DI) )}} \\ \hline
    \multirow{2}{*}{\textbf{Dataset}} & \multirow{2}{*}{\textbf{Models (\%)}} & \multirow{2}{*}{\textbf{EER (\%)}} & \multicolumn{2}{c}{\textbf{TMR (\%) @ FMR (\%) =}} \\ \cline{4-5}
     &  &  & \textbf{1 (\%)} & \textbf{0.1 (\%)} \\ \hline
     DI-P1 & Open-Iris & \textbf{4.86} & \textbf{90.52} & \textbf{75.57}\\ \hline
    DI-P1 & ViT-b & 12.53 & 51.44 & 20.98 \\ \hline
    DI-P1 & Resnet50 & 17.74 & 33.62 & 11.78 \\ \hline
    DI-P2 & Open-Iris & \textbf{8.00} & \textbf{43.10} & \textbf{10.63}\\ \hline
    DI-P2 & ViT-b & 18.10 & 42.82 & 22.13 \\ \hline
    DI-P2 & Resnet50 & 17.23 & 50.00 & 35.06 \\ \hline
    DI-P3 & Open-Iris & \textbf{9.16} & \textbf{94.94} & \textbf{44.83} \\ \hline
    DI-P3 & ViT-b & 17.39 & 55.17 & 33.05 \\ \hline
    DI-P3 & Resnet50 & 24.13 & 32.76 & 14.94 \\ \hline
    \end{tabular}    
    \label{table:table2}
\end{table*}
The results in Table \ref{table:table2}
 illustrate the performance of different iris recognition models in dark iris data sets captured using devices P1, P2, and P3. The Open-Iris model consistently outperforms the other models across all datasets, demonstrating its robustness in recognizing dark irises. Specifically, it achieves the lowest Equal Error Rates (EERs) of $4.86\%$, $8.00\%$, and $9.16\%$ for the P1, P2 and P3 data sets, respectively. Furthermore, Open-Iris maintains high True Match Rates (TMRs), with values exceeding $90\%$ at both the $1\% $and $0.1\%$ False Match Rate (FMR) thresholds, indicating strong reliability between different devices. 

In contrast, the ViT-b and ResNet-50 models show significantly lower performance, particularly in scenarios with dark irises. The ViT-b model exhibits higher EERs ranging from $12.53\%$ to $18.10\%$, and its TMRs are notably weaker, particularly at the $0.1\%$ FMR threshold, where TMRs drop as low as $22.13\%$ for the P2 dataset. ResNet-50 also struggles, with EERs as high as $24.13\%$ and TMRs that drop dramatically on the P1 dataset at the $0.1\%$ FMR threshold. These results highlight that, while Open-Iris remains effective across different datasets, both ViT-b and ResNet-50 are less reliable for dark iris recognition, especially in more challenging conditions, underscoring the importance of selecting models that can generalize well across varying iris pigmentation.
 \begin{figure}[!ht]
\centering
\includegraphics[scale=0.3]{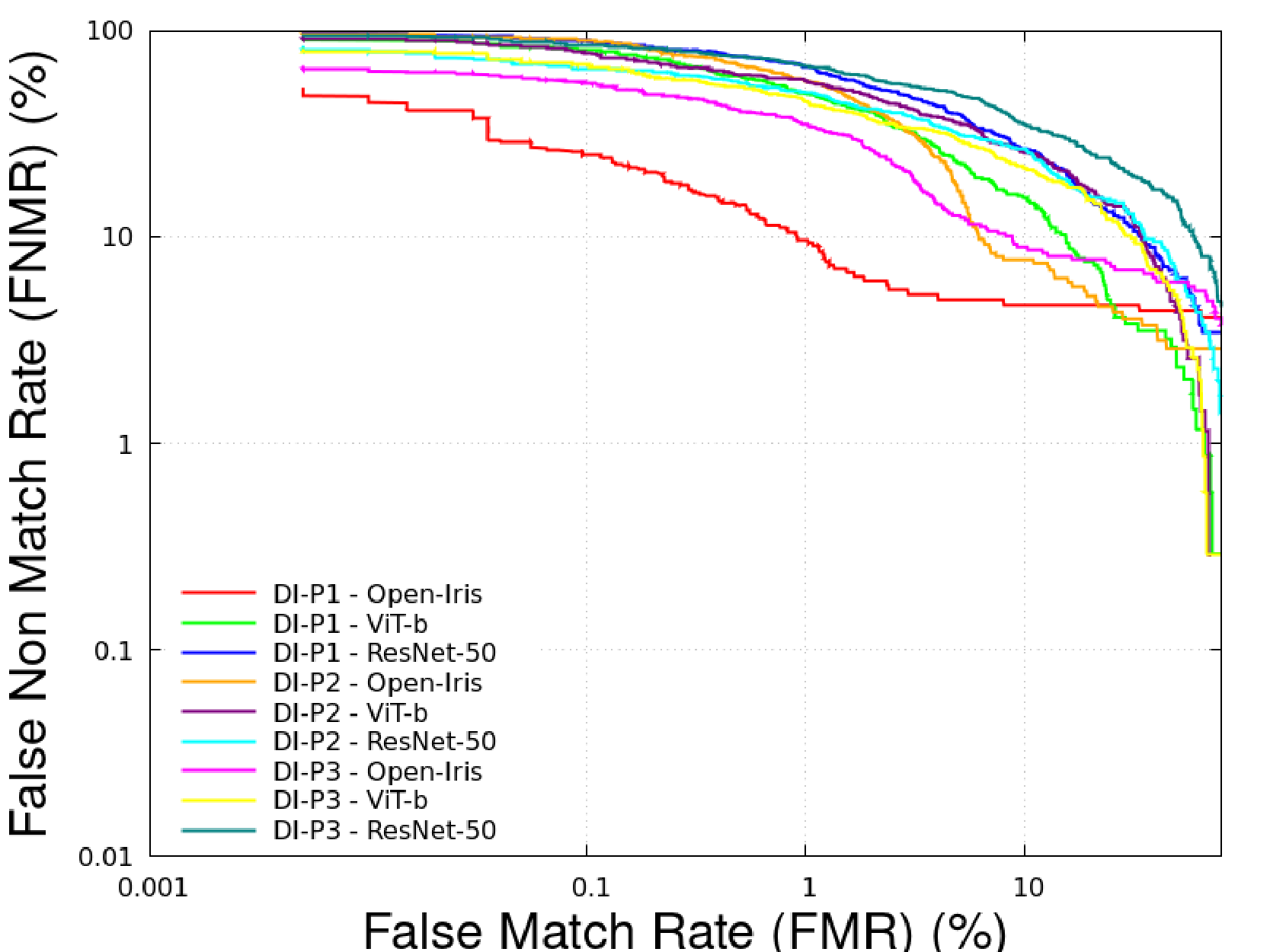}
\caption{ DET curve for comparative analysis among three different models. The X-axis indicates the false match rate and the y-axis indicates the false non-match rate of the Iris Pigmentation (Dark color) dataset.}
\label{fig:table2-fig}
\end{figure}

Comparing the results from Table \ref{table:table1} and Figure \ref{fig:table1-fig} (blue iris) and Table \ref{table:table2} and Figure \ref{fig:table2-fig} (dark iris), it is visible that the Open-Iris model consistently performs well across both iris pigmentation types, achieving low EERs and high TMRs. However, the performance of ViT-b and ResNet-50 significantly declines with dark irises, showing higher EERs and lower TMRs, particularly at lower FMR thresholds. Results suggests that iris pigmentation, particularly darker pigmentation, poses a greater challenge for these models compared to blue irises.

\subsection{Result: Demographic Fairness in Iris Recognition}
Addressing fairness in biometric systems( Open-Iris, ViT-b, and ResNet-50 explained in methodology section), particularly in iris recognition, is crucial to ensure equitable performance across diverse populations. System biases can arise due to varying factors such as differences in iris pigmentation, capture devices, and environmental conditions during image acquisition. 
Table \ref{table:table3} presents the fairness performance of the iris recognition system, specifically analyzing bias across different devices for the same iris color. The DPD\cite{kotwal2022fairness} and EoD\cite{kotwal2022fairness} metrics are used to measure this bias. In our experiment, DpD and EoD values are computed at the 0.1 threshold. The results indicate that blue irises (BI-P1 - B1-P2) exhibit minimal bias with low DPD $(7.18\%)$ and EoD $(7.76\%)$, suggesting a fair performance across devices.
\begin{table}[htbp]
    \centering
    \caption{Fairness Performance : Consistent Iris Color Across Different Devices}
    \begin{tabular}{ccc}
    \hline
    \textbf{Method} & \textbf{DPD($\%$)} & \textbf{EoD($\%$)} \\ \hline
    BI-P1 - BI-P2 & \textbf{7.18} & \textbf{7.76} \\ \hline
    DI-P1 - DI-P2 & 64.94 & 68.08 \\ \hline
    DI-P1 - DI-P3 & 30.74 & 35.04 \\ \hline
    DI-P2 - DI-P3 & 34.20 & 35.36 \\ \hline
    \end{tabular}    
    \label{table:table3}
\end{table}
 In contrast, dark irises display higher levels of bias, particularly between DI-P1 and DI-P2, where DPD $(64.94\%)$ and EoD $(68.08\%) $are significantly higher. This indicates that the recognition system is less consistent for dark irises when captured on different phones, highlighting a potential challenge in maintaining fairness across devices for varying iris pigmentation.
Table \ref{table:table4} provides an analysis of the fairness performance of the iris recognition system when comparing the same phone device across different iris colors that are presented using the DPD and EoD for two specific device comparisons: BI-P1 versus DI-P1 and BI-P2 versus DI-P2. The results reveal a notable disparity in fairness performance based on iris color within the same device. For the BI-P1 versus DI-P1 comparison, the DPD is $23.28\%$, indicating a relatively medium difference in treatment between blue and dark irises. The EoD is $27.90\%$, suggesting that the performance disparity in terms of True Positive Rate (TPR) and False Positive Rate (FPR) between the two iris colors is small.
\begin{table}[htbp]
    \centering
    \caption{Fairness Performance : Consistent Device with Varying Iris Colors}
    \begin{tabular}{ccc}
    \hline
    \textbf{Method} & \textbf{DPD($\%$)} & \textbf{EoD($\%$)} \\ \hline
    BI-P1 - DI-P1 & \textbf{23.28} & \textbf{27.90} \\ \hline
    BI-P2 - DI-P2 & 81.04 & 88.22  \\ \hline
    \end{tabular}    
    \label{table:table4}
\end{table}
In contrast, the BI-P2 versus DI-P2 comparison shows a higher DPD of $81.04\%$ and an EoD of $88.22\%$. This suggests significant fairness issues, with the system exhibiting considerable bias against one iris color when using the same device (P2). The higher DPD and EoD indicate that the recognition performance varies more substantially based on iris pigmentation in this case, leading to potential inequities in recognition accuracy. 
Table \ref{table:table5} examines the fairness performance of the iris recognition system when evaluating cross-phone device comparisons with different iris colors. This table provides the DPD and EoD across several device combinations, assessing how system bias varies between devices and iris pigmentation. The results reveal significant variability in fairness performance across different cross-device comparisons. For instance, the BI-P1 versus DI-P2 comparison shows a high DPD of $88.22\%$ and an EoD of $95.98\%$. This indicates a substantial bias, with considerable disparities in recognition accuracy and fairness between blue and dark irises when using different devices. 
\begin{table}[htbp]
    \centering
    \caption{Fairness Performance : Cross-Device and Cross-Iris Color Evaluation}
    \begin{tabular}{ccc}
    \hline
    \textbf{Method} & \textbf{DPD($\%$)} & \textbf{EoD($\%$)} \\ \hline
    BI-P1 - DI-P2 & 88.22 & 95.98 \\ \hline
    BI-P1 - DI-P3 & 54.02 & 62.94 \\ \hline
    BI-P2 - DI-P1 & \textbf{16.10} & \textbf{20.14} \\ \hline
    BI-P2 - DI-P3 & 46.84 & 55.18 \\ \hline
    \end{tabular}
    \label{table:table5}
\end{table}
Similarly, the BI-P1 versus DI-P3 comparison exhibits a DPD of $54.02\%$ and an EoD of $62.94\%$, suggesting another instance of notable bias across different devices.
In contrast, the BI-P2 versus DI-P1 comparison has a low DPD of $16.10\%$ and an EoD of $20.14\%$, indicating minimal bias and more consistent performance across different iris colors. However, the BI-P2 versus DI-P3 comparison shows a higher DPD of $46.84\%$ and an EoD of $55.18\%$, reflecting a moderate level of bias. These results highlight the pronounced variability in fairness performance across different devices and iris colors. While some device comparisons reveal significant bias, others demonstrate relatively equitable performance. 

\subsection{Discussion}
\begin{itemize}
    \item \textbf{Q1.} Does iris pigmentation color (Blue vs. Dark) impact the accuracy of the biometric system ?\\
    Yes, iris color significantly impacts the accuracy of the biometric system. The results of Tables \ref{table:table1} and \ref{table:table2} indicate that the recognition performance varies between the blue and dark irises. For example, Table \ref{table:table1} shows that the Open-Iris model achieved an EER of $0.24\%$ for blue irises captured by a P1, whereas it had an EER of $0.82\%$ for dark irises. Similarly, Table \ref{table:table2} highlights that dark irises generally exhibit higher error rates compared to blue irises. This variability is attributed to differences in how light and pigmentation affect the detection and recognition processes. \\
    \item \textbf{Q2.} Can the biometric system effectively generalize across varying iris colors, specifically between blue and dark eyes? \\
    The biometric system’s ability to generalize across varying iris colors is limited. As observed in Tables \ref{table:table1} and \ref{table:table2}, the system shows varying performance levels depending on the color of the iris and the device used. \\
\end{itemize}
\section{Conclusion}
\label{sec:5_conclusion}

This work explores the impact of iris pigmentation on biometric recognition, comparing the performance of blue and dark irises across various devices. Results indicate that recognition systems generally perform better on blue irises, showing lower Equal Error Rates (EER) and higher True Match Rates (TMR) compared to dark irises. Notably, the Open-Iris model demonstrated superior performance on blue irises, suggesting that iris pigmentation affects recognition accuracy. While incorporating a diverse range of iris colors enhances system generalization, performance is still influenced by factors such as device type and model architecture. A fairness analysis revealed biases, with varying levels of Differential Performance Disparity (DPD) and Equalized Odds Difference (EoD) depending on iris color and device, ultimately impacting both accuracy and fairness.

\vspace{-1cm} 
\bibliographystyle{plain}
\bibliography{ref.bib}{}
\end{document}